# AD-SAM: Adapting the Segment Anything Model for Semantic Segmentation in Autonomous Driving

Het Patel[a], Mario Camarena[b], Fatemeh Nazari[1,c], Evangelos Papalexakis[a], Mohamadhossein Noruzoliaee[c], Jia Chen[d]

[a] *Department of Computer Science, University of California Riverside, Riverside, CA 92521, USA*

[b] *Department of Computer Science, University of Texas Rio Grande Valley, Edinburg, TX 78539, USA*

[c] *Department of Civil Engineering, University of Texas Rio Grande Valley, Edinburg, TX 78539, USA*

[d] *Department of Electrical and Computer Engineering, University of California Riverside, Riverside, CA 92521, USA*

**Abstract:** This paper presents the Autonomous Driving Segment Anything Model (AD-SAM), a foundation-model adaptation framework for semantic segmentation in autonomous driving. AD-SAM combines a frozen Segment Anything Model (SAM) Vision Transformer (ViT-H) encoder with a trainable ResNet-50 encoder to integrate general visual representations with multi-scale, domain-specific spatial features. Features from the two encoders are integrated through deformable convolution and channel attention, followed by a multi-stage deformable decoder for semantic prediction. Training employs a hybrid objective combining Focal, Dice, Lovász-Softmax, and Surface losses. Experiments on Cityscapes and Berkeley DeepDrive 100K (BDD100K) show that AD-SAM outperforms SAM, Generalized SAM (G-SAM), and DeepLabV3 under a controlled training protocol. AD-SAM achieves 76.27% mIoU on Cityscapes and 64.74% on BDD100K, exceeding DeepLabV3 by 3.45 and 5.00 percentage points, respectively, with larger gains over the SAM-based baselines. Sample-size experiments reveal dataset-dependent behavior. In particular, AD-SAM performs strongly across training sizes on Cityscapes, while its advantage on the more heterogeneous BDD100K becomes pronounced with increased training data. When trained on Cityscapes and directly evaluated on BDD100K, AD-SAM achieves the highest cross-dataset retention (84.88%) among the evaluated models. AD-SAM also converges rapidly, while precomputed frozen SAM embeddings reduce training memory requirements. These findings demonstrate the potential of combining general foundation-model representations with domain-specific multi-scale features for accurate and robust autonomous-driving semantic segmentation.

# 1. Introduction

Autonomous driving (AD) systems represent a transformative leap in intelligent transportation, integrating artificial intelligence (AI), sensor fusion, and control algorithms to enable safe navigation of vehicles in complex and dynamic road environments with reduced or no human intervention. Among the core components of an AD system (i.e., perception, planning, and control), the perception module forms the foundation by interpreting sensor data to generate a real-time understanding of the surrounding road environment. This situational understanding, achieved through detecting, classifying, and localizing objects and terrain features, directly shapes the performance and reliability of downstream planning and control, especially under adverse conditions [1]. Within the perception stack, semantic segmentation assigns class labels to image pixels to generate dense, scene-level representations of the driving environment. Unlike object detection, which primarily identifies and localizes discrete entities with bounding boxes [2] and informs control tasks such as braking and steering, semantic segmentation provides fine-grained categorization of drivable and non-drivable areas (e.g., roads, sidewalks, and vegetation), static and dynamic obstacles (e.g., buildings, pedestrians, and vehicles), and navigational cues (e.g., traffic signs and lane markings). This pixel-level context supports planning tasks such as path delineation, lane-level localization, obstacle avoidance, and reasoning about interactions with vulnerable road users. Consequently, semantic segmentation is an important component of scene understanding [3], robust real-time perception [4,5], and trajectory generation [6] in AD systems.

Despite substantial advances in deep-learning-based semantic segmentation, its application to AD continues to face several challenges. First, supervised segmentation relies heavily on densely annotated imagery, for which pixel-level labeling is costly and labor-intensive. This dependence becomes particularly restrictive when representative training data are required across different cities, roadway configurations, traffic conditions, and environmental settings. Second, generalizability issues arise which involve limited robustness to unseen environments (e.g., new regions like different cities, and varying road layouts like highways and rural roads) and adverse weather or lighting (e.g., fog, rain, nighttime, and direct sunlight) as well as class imbalance favoring more common object classes. Third, high-resolution dense prediction imposes substantial computational requirements, creating an important trade-off between representational capacity, segmentation accuracy, and deployment efficiency [3,4,7]. These challenges motivate perception models that more effectively use limited labeled data while maintaining reliable representations across heterogeneous driving environments.

Recent advances in vision foundation models offer promising solutions to these challenges. Rather than learning visual representations exclusively from a task-specific driving dataset, foundation models leverage large-scale pretraining to acquire broadly reusable visual representations that can subsequently be adapted to downstream tasks [8]. In image segmentation, the Segment Anything Model (SAM) introduces a particularly influential foundation-model paradigm through large-scale pretraining and promptable, class-agnostic mask generation [9], subsequently extended to image and video segmentation by SAM 2 [10]. Such pretrained representations offer reduced dependence on learning visual features entirely from limited domain-specific datasets, which provide a stronger initialization for specialized segmentation tasks. However, foundation models cannot inherently resolve the challenges of AD perception. Their successful use depends on the effective adaptation of their general-purpose representations to the spatial, semantic, and computational requirements of the target domain [11–13].

This limitation is particularly relevant to SAM, which is designed for prompt-conditioned, class-agnostic mask generation rather than the prompt-free, closed-set pixel classification required for AD perception. Recent adaptations address this mismatch by incorporating task-specific information, including multi-scale convolutional features in SAMNet [14], progressive learning in APSAM [15], structural and edge priors in TOE-SAM [16], and vision-language fusion in EVF-SAM [17]. While these studies verify the adaptability of SAM, they primarily target specialized tasks such as salient-object detection, anomaly detection, scene-text segmentation, and referring-expression segmentation, leaving its adaptation to semantic scene understanding for AD comparatively underexplored.

Contemporary segmentation architectures also highlight the importance of combining contextual and spatial information. DDRNet [18] and PIDNet [19] employ complementary convolutional pathways to preserve multi-scale and spatial details, while SegFormer [20] uses hierarchical Transformer representations and ViT-Adapter [21] adapts pretrained Transformers to dense prediction through spatial priors and multi-scale interactions. These advances suggest an opportunity to combine broadly pretrained foundation-model representations with domain-specific hierarchical features. However, such integration for AD segmentation, particularly under limited labeled data and heterogeneous driving datasets, remains insufficiently explored.

To address these gaps, this study proposes **AD-SAM (Autonomous Driving Segment Anything Model)**, a SAM-based adaptation framework for prompt-free semantic segmentation of AD scenes. Rather than fine-tuning the SAM image encoder, AD-SAM retains the pretrained SAM ViT-H encoder as a frozen representation pathway and complements it with a trainable ResNet-50 encoder [22] that extracts hierarchical, domain-specific features at multiple spatial scales. Features from the two pathways are integrated through deformable convolutional fusion with channel attention, followed by a multi-stage deformable convolutional decoder for dense semantic prediction. A hybrid objective combining Focal Loss, Dice Loss, Lovász-Softmax, and Surface Loss is further employed to address complementary aspects of class imbalance, region overlap, semantic segmentation accuracy, and boundary quality.

The proposed framework is evaluated on the Cityscapes [23] and BDD100K [24], which are driving-scene datasets under progressively varying training-set sizes. The study therefore examines not only overall segmentation performance but also how the benefit of adapting pretrained foundation-model representations changes with the amount of available labeled data and across heterogeneous driving datasets. The principal contributions of this study are threefold: (i) a dual-encoder adaptation architecture that integrates frozen SAM representations with trainable multi-scale convolutional features for prompt-free, closed-set AD semantic segmentation; (ii) an adaptive fusion and decoding framework that combines deformable convolution and channel attention to integrate general-purpose and domain-specific representations, with multi-scale convolutional features reintroduced through the decoder; and (iii) a systematic evaluation of segmentation performance across varying training-set sizes and two heterogeneous driving datasets, providing insight into data sensitivity and cross-dataset performance robustness of SAM-based adaptation for AD perception. The proposed approach advances generalizable perception by enhancing robustness, accuracy, and label efficiency, which are critical factors for trustworthy AI in autonomous transportation systems.

The remainder of the paper is structured as follows. Section 2 reviews related work followed by Section 3 presenting the proposed AD-SAM framework. The experimental setup is described in Sections 4 with results discussed in Section 5. Finally, Section 6 concludes the paper with key findings, limitations, and future research directions.

# 2. Related Work

## 2.1. Vision Foundation Models for Semantic Segmentation

Vision foundation models introduce a new paradigm in computer vision by leveraging large-scale pretraining to learn general-purpose representations transferrable across visual domains and downstream tasks. These models shift computer vision from task-specific toward reusable representation learning with greater potential for generalization and adaptation [8]. This transition is particularly relevant to image segmentation, where conventional supervised approaches depend heavily on densely annotated data. Recent reviews highlight the potential of foundation models to combine broadly pretrained visual knowledge with the spatial information required for pixel-level prediction [13].

Among these models, the SAM has substantially influenced recent segmentation research. SAM performs promptable, class-agnostic mask generation with strong zero-shot capabilities [9] followed by SAM 2, which extends this paradigm to images and videos through memory-based temporal processing [10]. Subsequent research explores SAM's applications, prompting strategies, and adaptation to specialized domains [11,12]. More broadly, prompt-based and learned adaptation emerges as important mechanisms for transferring foundation-model knowledge to downstream visual tasks [25].

However, SAM is not inherently designed for closed-set semantic segmentation, which requires assigning predefined semantic classes to all pixels rather than generating class-agnostic masks from prompts. This mismatch motivates approaches that preserve SAM's pretrained representations while introducing task-specific features or adaptation mechanisms. For example, SAMNet incorporates a multi-scale convolutional adapter and cross-modal fusion for light-field salient object detection while largely freezing the SAM encoder [14]. APSAM employs adaptive progressive learning to specialize SAM for anomaly detection [15]. TOE-SAM introduces text-structure and edge priors through a lightweight adapter for scene-text segmentation while retaining a frozen SAM encoder [16], whereas EVF-SAM uses early vision-language fusion to adapt SAM for text-prompted referring-expression segmentation [17].

Overall, these studies verify the value of complementing pretrained SAM representations with task-specific information. However, existing adaptations primarily address salient-object detection, anomaly detection, scene-text segmentation, and referring-expression segmentation rather than prompt-free, closed-set semantic segmentation of complex driving scenes. AD perception additionally requires simultaneous recognition of heterogeneous road users and infrastructure across multiple spatial scales and under varying environmental and data conditions. Therefore, effectively integrating SAM's general-purpose representations with domain-specific, multi-scale features for prompt-free AD semantic segmentation remains an important research direction.

## 2.2.Semantic Segmentation for Autonomous Driving

Semantic segmentation is a fundamental component of AD perception, providing dense, pixel-level understanding of road surfaces, vehicles, pedestrians, cyclists, traffic infrastructure, vegetation, and other scene elements. Deep-learning approaches advance this task through convolutional, encoder-decoder, multi-scale, and attention-based architectures [26–28]. AD perception, however, imposes requirements beyond conventional benchmark segmentation accuracy. Models need to operate in dynamic and safety-critical environments characterized by considerable variation in road configuration, traffic composition, illumination, weather, object scale, and sensing conditions. Reliable recognition of comparatively small yet safety-relevant classes, such as pedestrians, riders, traffic signs, and traffic lights, need to coexist with accurate delineation of large structural regions such as roads, sidewalks, buildings, and vegetation. Muhammad et al. [29] identify perception reliability, environmental variability, computational efficiency, and safety as persistent challenges for deep-learning-based AD. These characteristics make robustness and generalizability particularly important considerations for semantic perception systems.

One direction for addressing environmental complexity is the integration of complementary sensing modalities. Feng et al. [30] review multimodal object-detection and semantic-segmentation methods that combine information from cameras, LiDAR, radar, and other sensors, highlighting their potential to improve perception robustness under challenging conditions. For image-based segmentation, research instead emphasizes on improvements in feature extraction and aggregation, including multi-scale representations and mechanisms for retaining spatial detail while capturing broader contextual information [27,28]. Across both directions, AD perception involves a persistent trade-off among segmentation accuracy, robustness, and computational requirements.

A related challenge is the dependence of supervised segmentation models on densely annotated driving imagery. Pixel-level annotation is expensive, while the visual characteristics represented by a particular dataset may cover only a subset of the conditions encountered in deployment. Differences in geographic context, road infrastructure, traffic composition, weather, illumination, and data-acquisition systems can therefore lead to substantial distributional variation across driving datasets and operating environments [29,30]. As a result, performance obtained from a single benchmark may not fully characterize a model's ability to maintain segmentation quality under different data conditions. These limitations motivate greater attention to both data efficiency and robustness across heterogeneous driving environments, particularly when developing visual perception systems intended for autonomous navigation.

## 2.3.Recent Architectures for Efficient and Robust Semantic Segmentation

Recent semantic segmentation research increasingly seeks to balance representation quality, preservation of spatial detail, and computational efficiency. Earlier deep segmentation architectures establish the effectiveness of convolutional feature hierarchies and encoder-decoder structures for dense prediction [26,28]. Subsequent architectures extend these principles through multi-scale contextual modeling, parallel-resolution processing, boundary-aware representations, and increasingly efficient decoding strategies. Such developments are particularly relevant to AD, where a segmentation model requires to simultaneously capture broad scene context and preserve fine spatial information for comparatively small or irregularly shaped road users and infrastructure.

Modern convolutional architectures increasingly address this challenge through explicit multi-resolution processing. DDRNet [18], for example, maintains parallel high- and low-resolution representations and repeatedly exchanges information between them, allowing detailed spatial information to be retained while incorporating contextual features. PIDNet [19] further develops this concept through separate yet interacting branches for detail, context, and boundary information. These approaches verify the continuing effectiveness of convolutional inductive biases for dense scene understanding while explicitly targeting the accuracy-efficiency trade-off required by real-time visual perception.

In parallel, Transformer-based architectures expand the representation capabilities available for semantic segmentation. SegFormer [20] combines a hierarchical Transformer encoder with a lightweight decoder, enabling multi-scale representation learning while capturing long-range contextual dependencies. Other approaches focus on adapting pretrained Vision Transformers to dense prediction rather than training task-specific Transformer architectures from the outset. ViT-Adapter [21], for example, introduces spatial priors and multi-scale feature interactions to adapt pretrained Transformer representations to dense prediction tasks. These developments show a broader methodological trend toward combining general pretrained representations with mechanisms that recover the spatial and multi-scale information required for pixel-level prediction.

The emergence of vision foundation models extends this progression further. In contrast to conventional convolutional neural network- and Transformer-based segmentation architectures, which are optimized specifically for dense semantic prediction, models such as SAM begin with representations learned from substantially broader visual data and objectives. This provides potentially richer and more transferable visual knowledge but also introduces a mismatch with closed-set semantic segmentation, particularly when interactive prompting is unavailable. Consequently, the adaptation of foundation models requires mechanisms capable of preserving their general visual knowledge while introducing the domain-specific spatial and semantic information required for the target environment.

The literature therefore reveals complementary strengths and limitations across task-specific segmentation architectures and general-purpose vision foundation models. Conventional convolutional neural network - and Transformer-based approaches provide effective multi-scale representations and class-specific dense prediction but remain dependent on domain-specific training data. Foundation models such as SAM provide broadly pretrained visual representations but are not inherently designed for prompt-free semantic labeling of complex driving scenes. Moreover, comparatively limited attention is given to integrating these complementary representations while systematically examining performance under reduced training data and heterogeneous AD datasets. This gap motivates AD-SAM, which combines the general-purpose representation capability of a frozen SAM encoder with trainable multi-scale convolutional features, adaptive feature fusion, and semantic decoding for AD perception. The framework is evaluated across varying training-set sizes and heterogeneous driving datasets to examine both segmentation performance, and sensitivity to the amount and characteristics of available training data.

# 3. Modeling Framework

This section introduces the overall AD-SAM architecture (Figure 1), followed by a detailed description of its components including the dual encoder, the deformable feature fusion module, the progressive decoder, and the hybrid loss. Throughout, the SAM image encoder serves as a frozen feature extractor meaning that no SAM parameters are updated. The trainable part of AD-SAM is a convolutional branch that runs in parallel with the frozen encoder, along with a lightweight fusion and decoding path that combines the two.

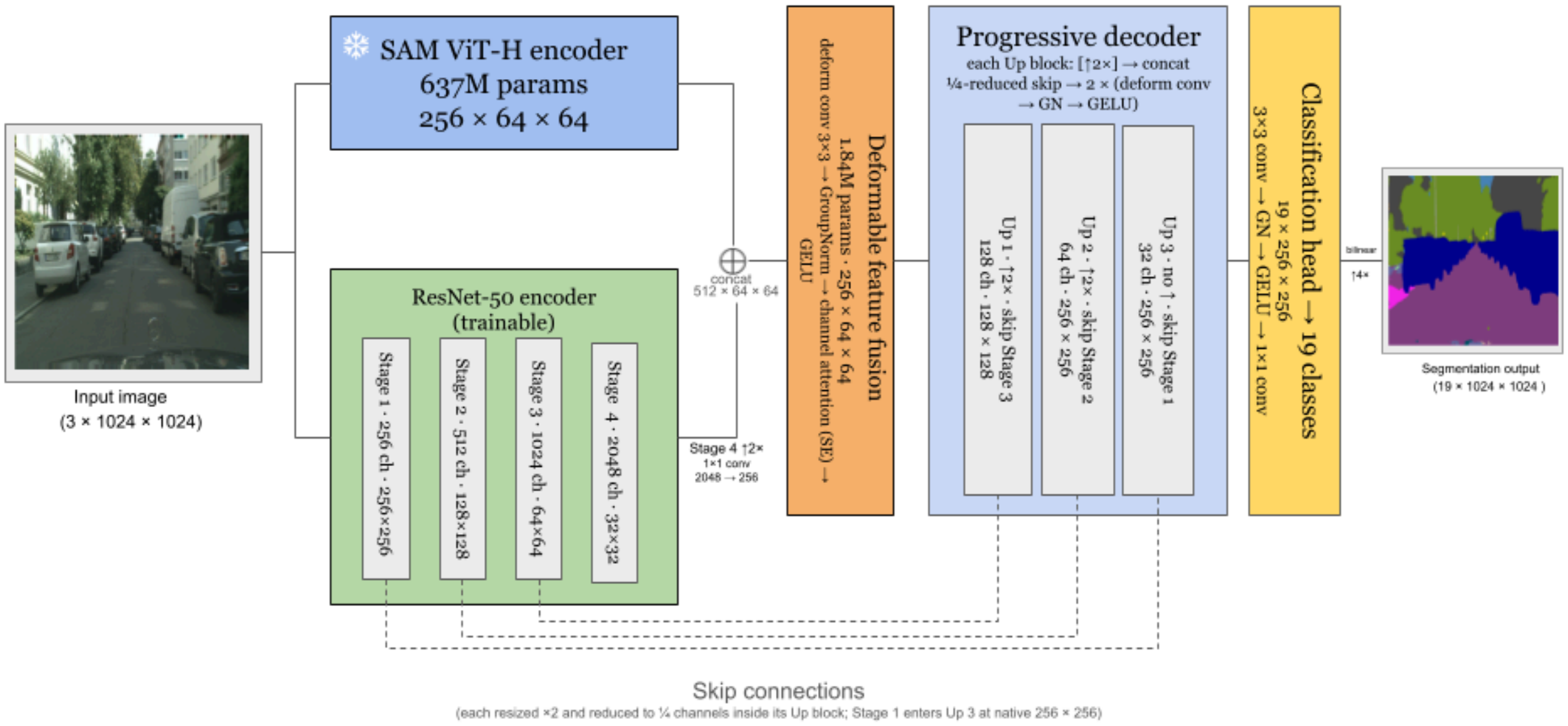

**Figure 1. Overarching framework of AD-SAM**

## 3.1.Dual-Encoder

The dual-encoder architecture leverages the complementary strengths of vision transformers (ViT-H in SAM) for global context and convolutional networks (ResNet-50) for local spatial detail and task-specific semantics.

**Frozen SAM encoder.** In SAM, the ViT-H image encoder [31] processes inputs at a resolution of 1024 × 1024, dividing the image into 16 × 16 patches and passing the resulting 64 × 64 grid of tokens through 32 transformer blocks with a learned absolute positional embedding. A two-layer convolution neck projects the 1280-dimensional tokens to 256 channels. This yields feature embeddings $F_{SAM} \in R^{(B\times256\times64\times64)}$, where $B$ is the batch size: one 256-dimensional descriptor for every 16 ×16 region of the input. The encoder receives the image in SAM's native normalization and remains frozen to preserve the representations learned from pretraining on the SA-1B dataset [9]. Since the encoder is deterministic for a given image, $F_{SAM}$ is computed once per image and stored. Also, the encoder is not executed again during training (Section 3.6).

**Trainable ResNet-50 encoder.** Concurrently, a ResNet-50 [22] backbone initialized with ImageNet weights which processes the ImageNet-normalized image and extracts hierarchical features at four scales through its residual stages, yielding:

- $F_{res1}$: **256 channels at stride 1/4 (256 × 256)**
- $F_{res2}$: **512 channels at stride 1/8 (128 × 128)**
- $F_{res3}$: **1024 channels at stride 1/16 (64 × 64)**

- $F_{res4}$: **2048 channels at stride 1/32 (32 × 32)**

The global average pooling and classification layers of the original network are removed. The deepest map $F_{res4}$ carries the most semantic content and enters the fusion module; $F_{res1}$ to $F_{res3}$ carry progressively finer spatial detail and enter the decoder as skip connections. The backbone is trained at one tenth of the learning rate of the newly initialized layers, and its batch-normalization statistics are kept fixed at their pretrained values, as is standard when training a pretrained backbone at a small batch size.

## 3.2.Deformable Feature Fusion

The two streams are combined at the 64 × 64 resolution of $F_{SAM}$. The ResNet map $F_{res4}$ is bilinearly upsampled from 32 × 32 to 64 × 64 and projected by a 1 × 1 convolution to SAM's 256-dimensional space, giving $F'_{res4} \in R^{(B\times256\times64\times64)}$. The two maps are concatenated along the channel dimension to form $F_{concat} \in R^{(B\times512\times64\times64)}$. Feature fusion then uses modulated deformable convolution [32,33], which adaptively learns spatial sampling patterns. The deformable fusion computes offset fields $\Delta_p \in R^{2k}$ and modulation masks $m \in R^k$ from $F_{concat}$ itself through two 3 × 3 auxiliary convolutions, where $k = 9$ for 3 × 3 kernels. The deformable operation is defined in Eq. (1), which transforms the input features $x(\cdot)$ by sampling at shifted locations $\left(p_0 + p_k + \Delta_{p_k}\right)$ and applying kernel weights $w_k$ and modulation terms $m_k$ to produce the fused feature $y(\cdot)$ at location $p_0$.

When sampling at fractional coordinates, we apply bilinear interpolation so that each output location pulls values from its selected positions. This allows the kernel to better match the transformer's patch-grid features, built from 16 × 16 patches that do not follow object boundaries, with the convolutional features from the ResNet branch. The offset and modulation convolutions are initialized to zero, so the layer starts as a regular convolution and deviates only if the data indicates it should. After the convolution step, group normalization normalizes the 256-channel fused map, and channel attention [34] recalibrates it. The attention module processes the fused features through parallel average-pooling and max-pooling paths, followed by a shared multi-layer perceptron with two fully connected layers and a reduction ratio of $r = 16$. The resulting attention vector $a$, shown in Eq. (2), adaptively re-weights feature channels. The re-weighted map passes through a GELU activation to produce the fused representation $F_{fused} \in R^{(B\times256\times64\times64)}$. Attention is applied after normalization so that the per-channel scaling it produces is not canceled by a subsequent normalization step. The fusion module contains 1.84M parameters.

$$y(p_0) = \sum_{k=1}^{K} w_k \cdot x(p_0 + p_k + \Delta p_k) \cdot m_k \quad (1)$$

$$a = \sigma\left(FC_2\left(ReLU\left(FC_1(avg(\bullet))\right)\right) + FC_2\left(ReLU\left(FC_1(max(\bullet))\right)\right)\right) \quad (2)$$

### 3.3. Progressive Decoder

The decoder takes fused features $F_{fused}$ as input and restores the spatial resolution via three sequential upsampling blocks, denoted $Up\,1$ to $Up\,3$ in Figure 1, each of which reintroduces one ResNet feature map as a skip connection. Every block performs the same sequence of operations: (i) the incoming decoder map is bilinearly upsampled by a factor of two; (ii) the corresponding ResNet skip map is resized to the same resolution if necessary and reduced to one quarter of its channels by 1 × 1 convolution, group normalization, and GELU; (iii) the two are concatenated along channels; and (iv) two modulated deformable convolutions of the form of Eq. (1), each followed by group normalization and GELU, produce the block output. The three blocks are:

- **$Up\,1$: $F_{fused}$ (256 ch, 64 × 64) upsampled to 128 × 128, fused with $F_{res3}$ (1024 → 256 ch) → 128 ch at 128 × 128**
- **$Up\,2$: 128 ch upsampled to 256 × 256, fused with $F_{res2}$ (512 → 128 ch) → 64 ch at 256 × 256**
- **$Up\,3$: 64 ch at 256 × 256 (no upsampling), fused with $F_{res1}$ (256 → 64 ch) at its native resolution → 32 ch at 256 × 256**

The final block operates at stride 4 without additional upsampling, so that the finest ResNet features are concatenated without resampling. All deformable convolutions in the decoder therefore run at 128 × 128 or 256 × 256, which keeps the cost of the deformable sampling operation modest. The 1 × 1 reduction applied to each skip serves two purposes. One is preventing the high-channel ResNet maps from dominating the concatenation, and the other is allowing the decoder dynamically learn which ResNet channels are informative at each resolution. The decoder and head contain 1.6M parameters.

A segmentation head produces final class predictions which are a 3 × 3 convolution with group normalization and GELU, followed by a 1 × 1 convolution that projects the 32-dimensional features to 19 semantic classes. The head runs at 256 × 256, and the resulting logits are bilinearly upsampled by a factor of 4 × to 1024 × 1024 input resolution, following common practice in encoder-decoder segmentation networks. The predicted labels at each pixel are the argmax over the 19 logits.

### 3.4. Hybrid Loss

The semantic class predictions are optimized using a composite loss formulated as a linear combination of four complementary loss terms (Eq. (3)). Focal loss [35] with $\alpha = 0.25$ and $\gamma = 2$ addresses class imbalance inherent in driving scenes. Dice loss [36] directly optimizes region overlap, while Lovász-Softmax loss [37] provides a smooth surrogate for discrete Intersection over Union (IoU) optimization. Surface loss [38] enhances boundary accuracy by weighting errors using distance transforms.

$$L_{total} = 0.4L_{focal} + 0.3L_{dice} + 0.2L_{lovász} + 0.1L_{surface} \tag{3}$$

### 3.5.Parameter Budget

The trainable parameters of each compared model are summarized in Table 1 in Section 4.3. For AD-SAM, the SAM ViT-H encoder contributes 637.0M frozen parameters. The trainable components are the ResNet-50 backbone (23.5M), the deformable fusion module (1.84M), and the decoder and head (1.6M), for a total of 26.9M trainable parameters, or roughly 4% of the 664.0M-parameter model. Approximately 90% of the trainable capacity therefore resides in the pretrained convolutional backbone rather than in newly initialized layers.

### 3.6.Training with Pre-computed Embeddings

Since no SAM encoder parameters are updated, $F_{SAM}$ depends only on the input image. AD-SAM therefore computes $F_{SAM}$ once for every training and validation image, and once for its horizontally flipped copy to support flip augmentation, and stores the results. During training, the encoder is never executed, and the stored tensor replaces the forward pass. As a result, the per-iteration training cost drops to that of the 26.9M-parameter trainable branch on a 1024 × 1024 input. Pre-computing the embeddings reduces per-epoch training time by roughly 40% and halves peak GPU memory (7.5 GB versus 15.2 GB at batch size 2) relative to executing the encoder online. This means that adapting AD-SAM to a new dataset requires one pass of the encoder over that dataset followed by training only the trainable branch which takes hours for the subset sizes studied here and one to two days at full scale on a single GPU. The stored embeddings are numerically identical to the online forward pass to within half-precision rounding.

This design also fixes the augmentation protocol, namely, geometric transformations that change the image content presented to the encoder, such as random scaling and cropping, cannot be applied to stored embeddings. We therefore use horizontal flipping as the only augmentation for AD-SAM, SAM, and DeepLabV3; G-SAM keeps its published augmentation (random 512 × 512 crops, rotation, and flip), since crop-based training is the method itself. Running the encoder online would remove this restriction and allow richer augmentation, which may yield further gains at higher cost. We deliberately take the other side of this trade-off so that AD-SAM leverages the features of a large foundation encoder while keeping adaptation to a new dataset cheap enough to run on a single GPU, and the pre-computed pipeline makes that possible. This choice does not change the architecture, and the same model can be trained online with the encoder when the added cost is acceptable.

# 4. Experimental Setup

## 4.1. Datasets and Evaluation Metrics

Experiments are conducted on two established AD semantic segmentation benchmarks — Cityscapes and BDD100K — which differ substantially in scale and visual diversity. Cityscapes [23] contains high-resolution urban street-scene images collected across 50 cities. The dataset provides 5,000 images with fine pixel-level annotations, divided into 2,975 training, 500 validation, and 1,525 test images. An additional 20,000 images with coarse annotations are available but are not used in this study. The experiments use the 19 semantic classes adopted by the standard Cityscapes evaluation protocol. BDD100K (Berkeley DeepDrive 100K by Yu et al. (2020)) is a large-scale driving dataset covering diverse geographic, weather, illumination, and traffic conditions. Its semantic segmentation subset contains 10,000 pixel-level annotated keyframes, comprising 7,000 training, 1,000 validation, and 2,000 test images with 19 semantic classes. All 7,000 training images are used in the full-data experiments. Compared with Cityscapes, BDD100K provides a more heterogeneous driving environment and therefore complements Cityscapes for evaluating model behavior across different data regimes and domains.

Segmentation performance is evaluated using IoU and mean Intersection over Union (mIoU). For a semantic class, IoU is defined by Eq. (4), where *TP*, *FP*, and *FN* denote the numbers of true-positive, false-positive, and false-negative pixels, respectively. Overall segmentation performance is measured by averaging IoU across the $C$ semantic classes (Eq. (5)). IoU is used for the class-level analysis, while mIoU serves as the primary metric for overall model comparison, sample-size sensitivity, and cross-dataset evaluation.

$$IoU = \frac{TP}{TP+FP+FN} \tag{4}$$

$$mIoU = \frac{1}{C}\sum_{i=1}^{C} IoU_i \tag{5}$$

## 4.2. Implementation and Training Setup

AD-SAM is implemented in PyTorch 2.7.0+cu118 with Python 3.12.10 and torchvision 0.20.0 under Ubuntu 22.04. Training is performed on a single NVIDIA RTX A6000 GPU. Automatic mixed-precision training is implemented using `torch.cuda.amp.autocast` and `GradScaler` to reduce memory consumption and accelerate computation [39].

Input images are resized to 1024 × 1024 pixels and normalized using ImageNet statistics (mean = [0.485, 0.456, 0.406], standard deviation = [0.229, 0.224, 0.225]). The model is optimized using AdamW with a base learning rate of $2 \times 10^{-4}$ and weight decay of $5 \times 10^{-4}$, together with cosine annealing without warm-up. Training is conducted for 100 epochs with a batch size of 2, and gradient clipping is not applied. The pretrained SAM ViT-H encoder remains frozen throughout training, whereas the ResNet-50 encoder, multi-scale fusion module, decoder, and segmentation head are optimized. Learning-rate multipliers of 0.1 and 1.0 are applied to the ResNet-50 and fusion/decoder components, respectively. Since the ViT-H encoder is frozen, SAM image embeddings are precomputed and reused during optimization rather than recomputed at every epoch. This strategy substantially reduces GPU memory requirements for the trainable AD-SAM branch. At a batch size of 2, peak GPU memory decreases from 15.2 GB with online SAM encoding to 7.5 GB when stored embeddings are used. The resulting trainable portion of AD-SAM contains 26.9 million parameters.

## 4.3.Baseline Model and Experimental Protocol

AD-SAM is compared with three reference models including the original SAM adapted for semantic segmentation, Generalized SAM (G-SAM) [40], and the convolutional neural networks-based DeepLabV3 [41]. These baselines provide complementary comparisons with a foundation-model representation, an existing SAM-based segmentation architecture, and a conventional fully trainable semantic segmentation network, respectively.

To support controlled comparison, the models are trained for 100 epochs with a training batch size of 2 and evaluated on the same validation partitions. Unless required by the architecture, inputs are standardized to 1024 × 1024 pixels and single-scale inference is used without test-time augmentation. G-SAM retains its architecture-specific input configuration of 512 × 1024. The resulting comparisons therefore characterize performance under a common, constrained computational budget rather than reproducing the individually optimized training recipes used to obtain each model's published benchmark results. Differences between the present protocol and published baseline configurations are discussed in Section 5.1.

Model complexity and inference characteristics are summarized in Table 1. AD-SAM has the largest total parameter count due to incorporating the frozen ViT-H encoder together with ResNet-50. However, only 26.9 M of its 664.0 M parameters are trainable. This distinction is important because total model size reflects inference complexity, whereas trainable parameters more directly characterize the optimization burden during adaptation. AD-SAM consequently trades increased inference computation for a substantially smaller trainable parameter set than G-SAM and DeepLabV3. Training-time scalability, including the one-time embedding precomputation for AD-SAM and SAM, is examined separately in Figure 7.

**Table 1. Model complexity and inference cost at each model's native input size (batch 1, FP16, RTX A6000).**

| Configuration | DeepLabV3 | SAM | G-SAM | AD-SAM (ours) |
|---|---|---|---|---|
| Backbone | ResNet-101 | ViT-H | ViT-B | ViT-H + ResNet-50 |
| Input resolution | 1024 × 1024 | 1024 × 1024 | 512 × 1024 | 1024 × 1024 |
| Parameters (M) | 61.0 | 643.2 | 160.0 | 664.0 |
| Trainable parameters (M) | 61.0 | 6.2 | 73.5 | 26.9 |

| Configuration | DeepLabV3 | SAM | G-SAM | AD-SAM (ours) |
|---|---|---|---|---|
| GFLOPs | 2,007 | 5,965 | 2,412 | 6,242 |
| Latency (ms) | 43.3 | 171.8 | 77.5 | 200.3 |
| FPS | 23.1 | 5.8 | 12.9 | 5.0 |
| Train/validation batch size | 2 / 2 | 2 / 2 | 2 / 4 | 2 / 2 |

# 5. Results and Discussion

## 5.1. Baseline Protocol and Scope of the Comparison

All models are evaluated under a controlled protocol using identical training subsets, 100 epochs, a batch size of 2, and the same validation images at 1024 × 1024 on a single GPU. The protocol is intentionally more constrained than the original baseline configurations. DeepLabV3, for example, reports 77.2% mIoU on Cityscapes using random scaling and cropping, a batch size of 16, and native-resolution evaluation [41], while the released G-SAM implementation reports 74.1% using a batch size of 8 over 200 epochs [40]. Under the present 100-epoch protocol, G-SAM achieves 67.36%, with its validation mIoU continuing to increase gradually near the end of training (Figure 4). The results should therefore be interpreted as a controlled comparison under a fixed computational budget rather than a ranking against published benchmark performance.

The protocol also reflects the resource-efficient adaptation strategy of AD-SAM. As the SAM encoder is frozen, its embeddings are precomputed and reused during training. This reduces peak GPU memory from 15.2 GB with online encoding to 7.5 GB for training the 26.9M-parameter trainable branch at a batch size of 2, enabling training within an 8 GB GPU memory budget. This efficiency entails a trade-off in that precomputed embeddings restrict online geometric augmentation. Accordingly, the comparison emphasizes model performance under standardized training and evaluation constraints rather than individually optimized training recipes. As shown subsequently, this setting enables assessment of whether AD-SAM can effectively exploit a large pretrained foundation encoder while limiting the memory and optimization requirements of downstream adaptation.

## 5.2. Overall Segmentation Performance

Table 2 compares the full-data segmentation performance of AD-SAM with DeepLabV3, SAM, and G-SAM. AD-SAM achieves the highest mIoU on both datasets, reaching 76.27% on Cityscapes and 64.74% on BDD100K. On Cityscapes, it improves upon DeepLabV3, G-SAM, and SAM by 3.45, 8.91, and 15.63 percentage points, respectively. The advantage is more pronounced on BDD100K, where AD-SAM exceeds DeepLabV3 by 5.00 points, SAM by 15.46 points, and G-SAM by 16.68 points. The stronger relative performance on BDD100K suggests that the integration of pretrained SAM representations with trainable multi-scale convolutional features is particularly beneficial in the more heterogeneous visual conditions represented by this dataset. Overall, these results verify the effectiveness of AD-SAM under the standardized training protocol.

**Table 2. Semantic segmentation performance (mIoU in %) across models and datasets**

| Dataset | Sample Size | DeepLabV3 | SAM | G-SAM | AD-SAM |
|---|---|---|---|---|---|
| Cityscapes | 2,975 | 72.82 | 60.64 | 67.36 | **76.27** |
| BDD100K | 7,000 | 59.74 | 49.28 | 48.06 | **64.74** |

## 5.3.Data Regime and Sample Size Sensitivity

Figure 2 and Figure 3 examine model performance across various training-set sizes. On Cityscapes, AD-SAM outperforms DeepLabV3 across all data regimes, with an advantage of approximately 2.9 mIoU points at 100 images and 3.5 points at the full 2,975 images (Figure 2(a)). Its advantage over SAM is largest under limited data and decreases as the sample size grows, whereas the gain over G-SAM increases toward the full-data setting (Figure 3(a)). These results indicate that AD-SAM remains competitive across data regimes while benefiting from additional training data.

A different pattern emerges on the more heterogeneous BDD100K dataset. DeepLabV3 performs better at 100, 500, and 1,000 training images, whereas AD-SAM surpasses it at the full 7,000-image training set by 5.0 mIoU points (Figure 2(b)). This crossover is consistent with differences in model initialization. AD-SAM retains pretrained SAM representations but must learn its trainable fusion, decoder, and classification components from the target data. Consequently, limited samples may be insufficient to learn these components across the broader visual variability of BDD100K. As the training set expands, AD-SAM increasingly benefits from this combination of pretrained representations and domain-specific feature learning, producing its largest absolute advantage at full data availability.

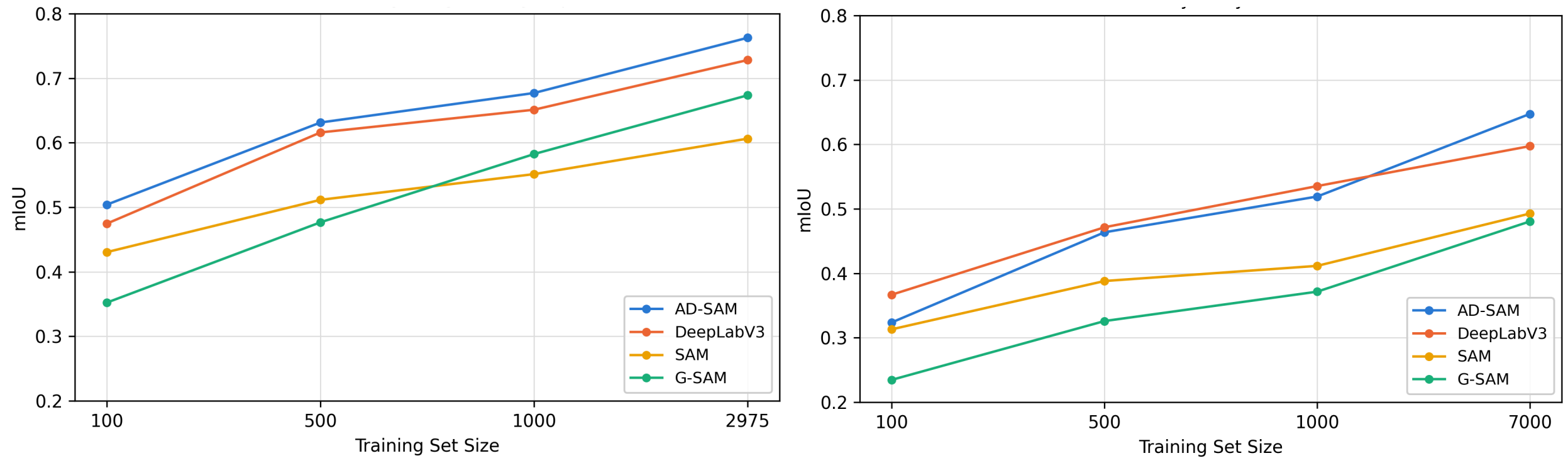

**(a) Cityscapes (full dataset)** **(b) BDD100K (full dataset)**

**Figure 2. Sensitivity analysis of semantic segmentation performance (mIoU) across varying training set sizes**

The relative gains in Figure 3 reinforce this dataset-dependent behavior. AD-SAM maintains substantial positive gains over SAM on both datasets, while its advantage over G-SAM generally becomes more pronounced with increasing sample size. On BDD100K, the relative gain over DeepLabV3 changes from negative under sparse training data to positive with the full dataset. Thus, the results do not indicate uniform superiority under extreme data scarcity. Rather, they show that the benefit of AD-SAM depends on both data availability and dataset diversity, with its advantage becoming stronger when sufficient observations are available to learn the task-specific components in heterogeneous driving environments.

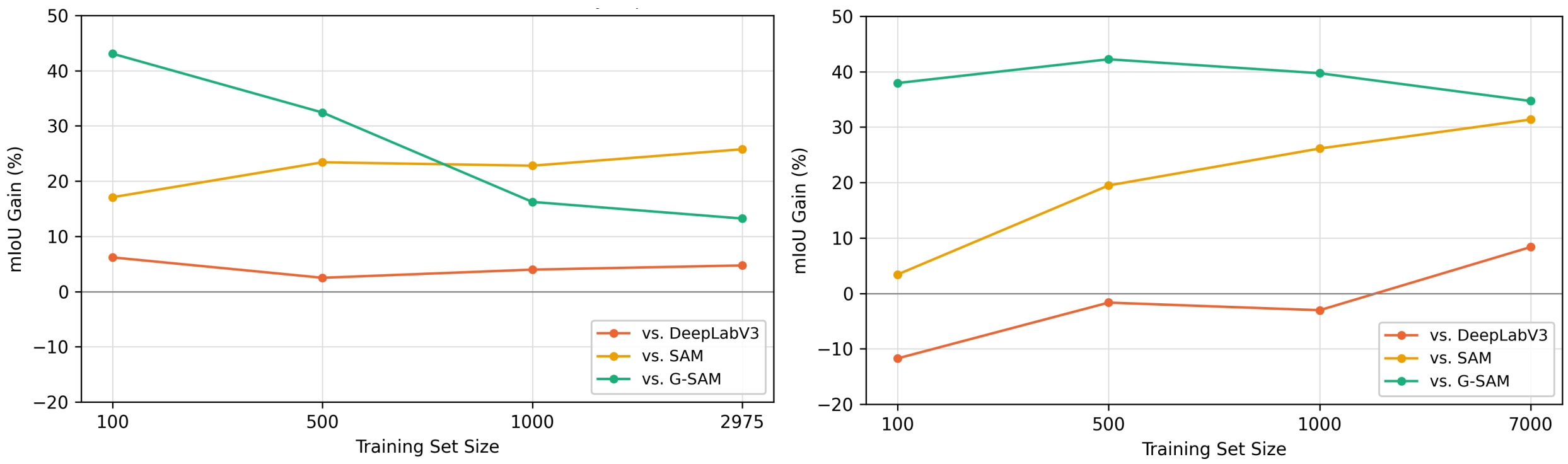

**(a) Cityscapes (full dataset)** **(b) BDD100K (full dataset)**

**Figure 3. Relative mIoU gains of AD-SAM over baseline models across training-set sizes**

## 5.4.Cross-Dataset Performance Robustness

Cross-dataset generalization is evaluated by training each model on Cityscapes and directly testing it on BDD100K without target-domain retraining. To account for differences in in-domain performance, cross-dataset retention is defined as the ratio of BDD100K transfer performance to Cityscapes in-domain performance, as written in Eq (6).

$$Retention = \frac{mIoU_{BDD100K}}{mIoU_{Cityscapes}} \tag{6}$$

As shown in Table 3, AD-SAM achieves the highest retention (84.88%), followed by DeepLabV3 (82.04%), SAM (81.27%), and G-SAM (71.35%). Thus, AD-SAM experiences the smallest proportional performance degradation when transferred from Cityscapes to BDD100K. This result is particularly relevant given the differences in visual and environmental conditions between the two datasets and suggests that combining frozen SAM representations with trainable domain-specific features improves robustness to dataset shift. Together with the sample-size analysis in Section 5.3, the findings indicate that AD-SAM benefits from both increased target-domain data and broadly pretrained representations, while maintaining comparatively strong performance when evaluated outside its training domain.

**Table 3. Cross-dataset performance retention (Cityscapes → BDD100K)**

| Model | Retention (%) |
|---|---|
| DeepLabV3 | 82.04 |
| SAM | 81.27 |
| G-SAM | 71.35 |
| **AD-SAM (ours)** | **84.88** |

## 5.5.Convergence and Computational Scalability

The validation trajectories in Figure 4 show distinct learning dynamics across models. AD-SAM converges rapidly on both datasets and maintains the highest validation mIoU through most of training. On Cityscapes, performance increases sharply during the early epochs before stabilizing near its final level, while G-SAM improves more gradually. A similar pattern is observed on BDD100K, where AD-SAM reaches a relatively stable high-performance region earlier than G-SAM. Notably, G-SAM remains on a gradual upward trajectory near epoch 100, indicating that its performance remains sensitive to the fixed training budget.

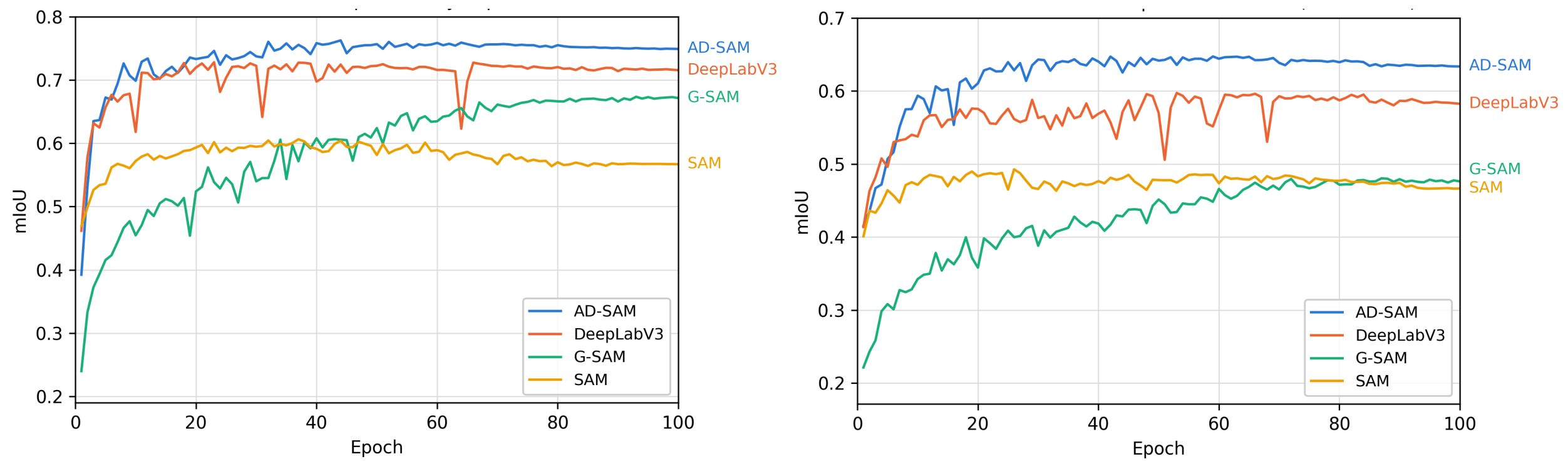

**(a) Cityscapes (full dataset)** **(b) BDD100K (full dataset)**

**Figure 4. Validation mIoU convergence of the evaluated models using the full training sets**

Figure 5 complements these results by showing validation-loss trajectories. Since AD-SAM and SAM use the hybrid loss, whereas G-SAM and DeepLabV3 use cross-entropy loss, absolute loss magnitudes are not directly comparable across loss formulations. Within the hybrid-loss models, AD-SAM exhibits a rapid initial reduction and earlier stabilization than SAM on both datasets, although some late-epoch variation is evident on BDD100K. The cross-entropy curves similarly show rapid initial convergence for DeepLabV3 and more gradual improvement for G-SAM. These results together with Figure 4 indicate that AD-SAM reaches strong validation performance relatively early within the 100-epoch training budget.

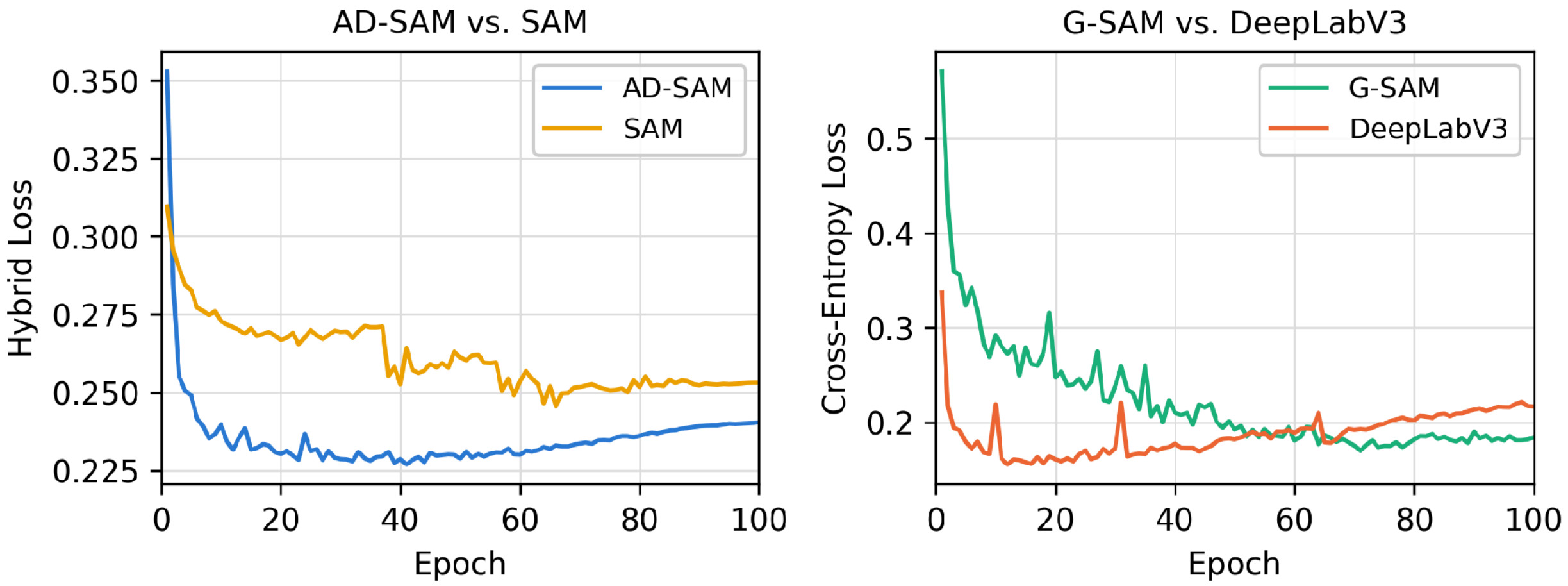

**(a) Cityscapes (full dataset)**

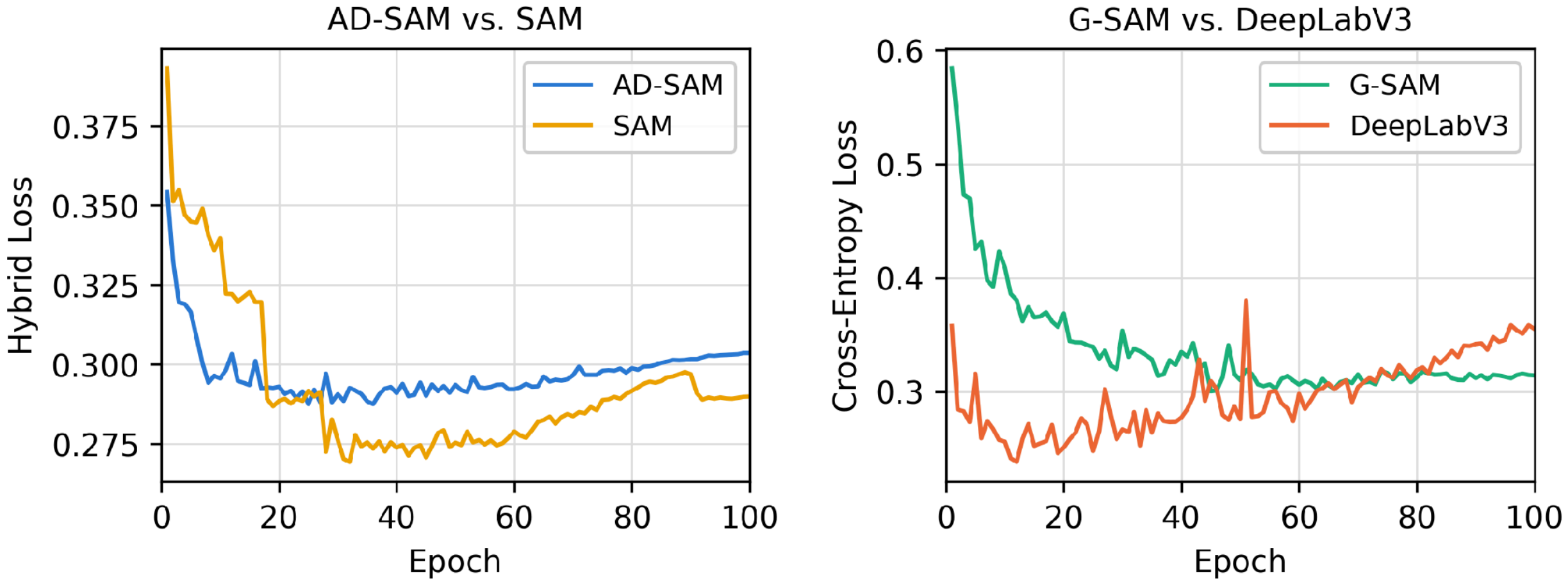

**(b) BDD100K (full dataset)**

**Figure 5. Validation loss convergence of the evaluated models on the full training sets**

The effect of data availability on learning dynamics is further illustrated in Figure 6. Across both datasets, increasing the training-set size generally raises the convergence level of the SAM-based models, although the magnitude of improvement varies across architectures and data regimes. AD-SAM exhibits particularly clear gains at medium and large sample sizes, with the full-data configurations reaching the highest validation mIoU on both datasets. The separation among BDD100K data regimes further supports the sample-size results in Section 5.3, showing that AD-SAM benefits substantially from additional observations in a heterogeneous driving environment.

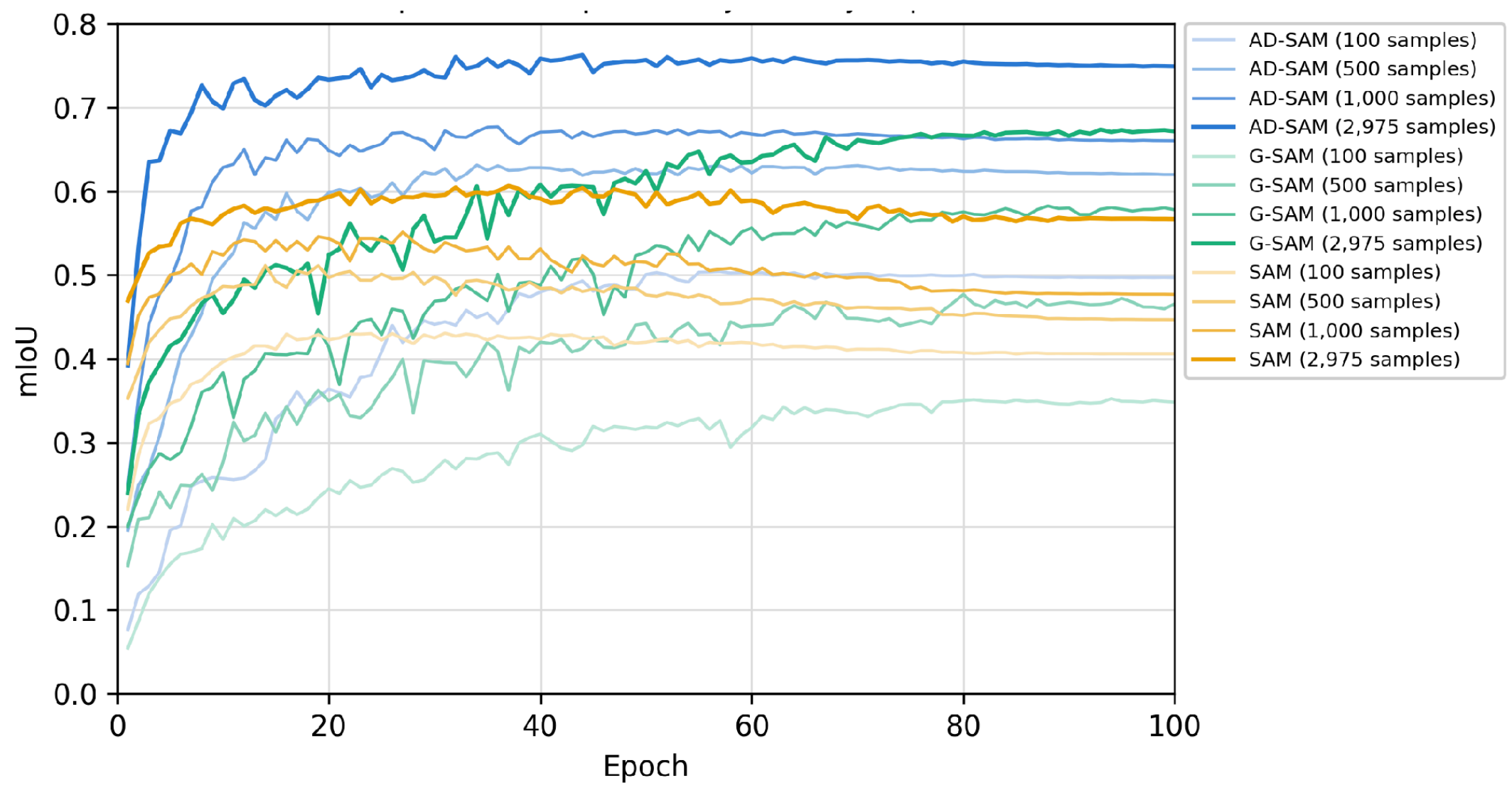

**(a) Cityscapes (full dataset)**

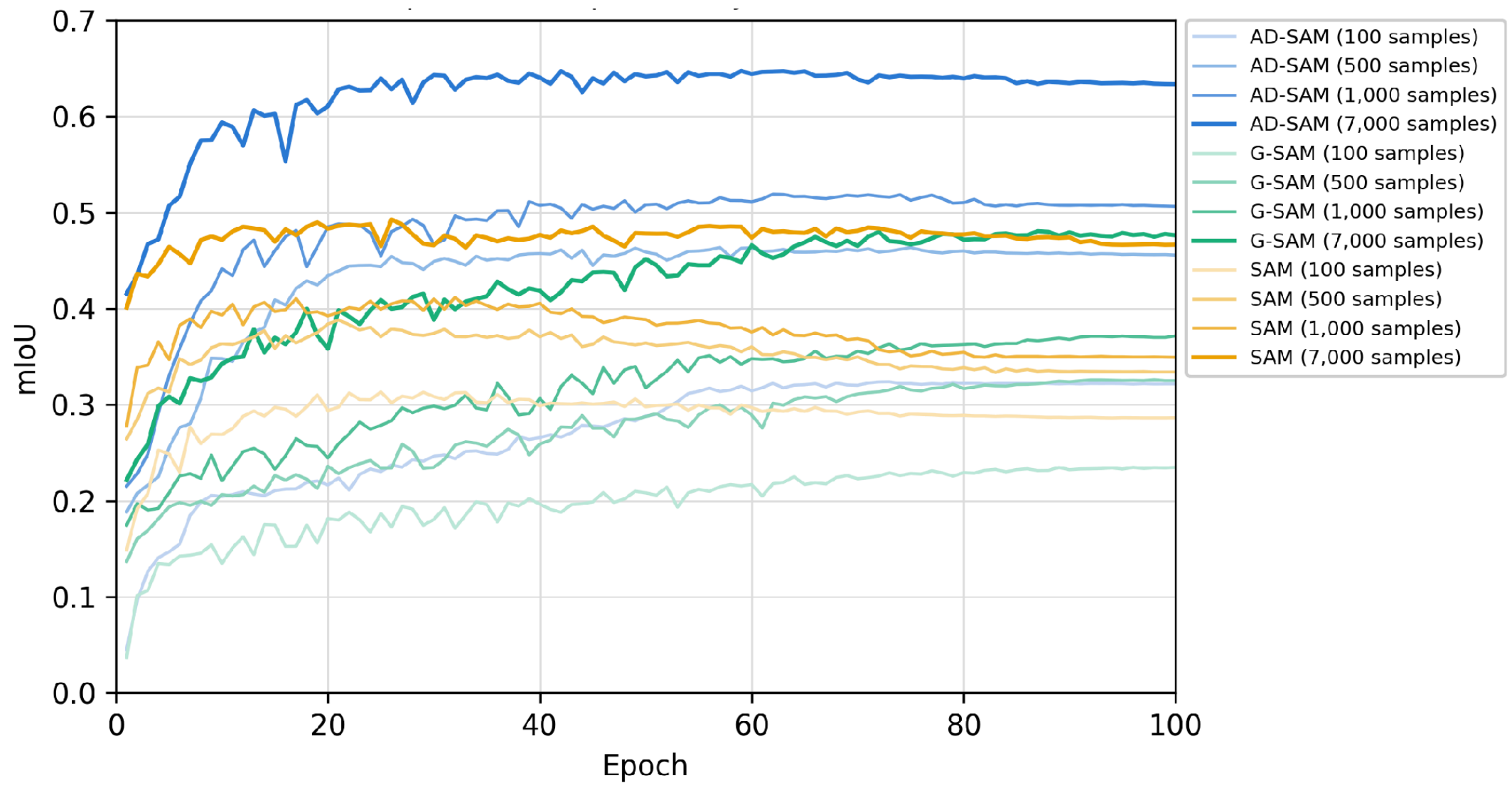

**(b) BDD100K (full dataset)**

**Figure 6. Validation mIoU trajectories of SAM-based models across training-set sizes**

Computational scalability presents a different trade-off (Figure 7). DeepLabV3 requires the least training time, whereas AD-SAM and G-SAM incur substantially higher costs as the dataset grows. At full scale, AD-SAM requires approximately 1,424 min on Cityscapes and 3,258 min on BDD100K, including the one-time SAM embedding precomputation, compared with approximately 1,421 and 3,198 min for G-SAM, respectively. Thus, AD-SAM does not provide the lowest training cost. Rather, its advantage lies in achieving higher full-data accuracy at a training cost comparable to G-SAM. Precomputing the frozen SAM representations shifts part of this cost to a one-time embedding stage while reducing the memory required during subsequent optimization, as discussed in Section 5.1. Overall, the results characterize AD-SAM as an accuracy-computation trade-off rather than a computationally minimal architecture.

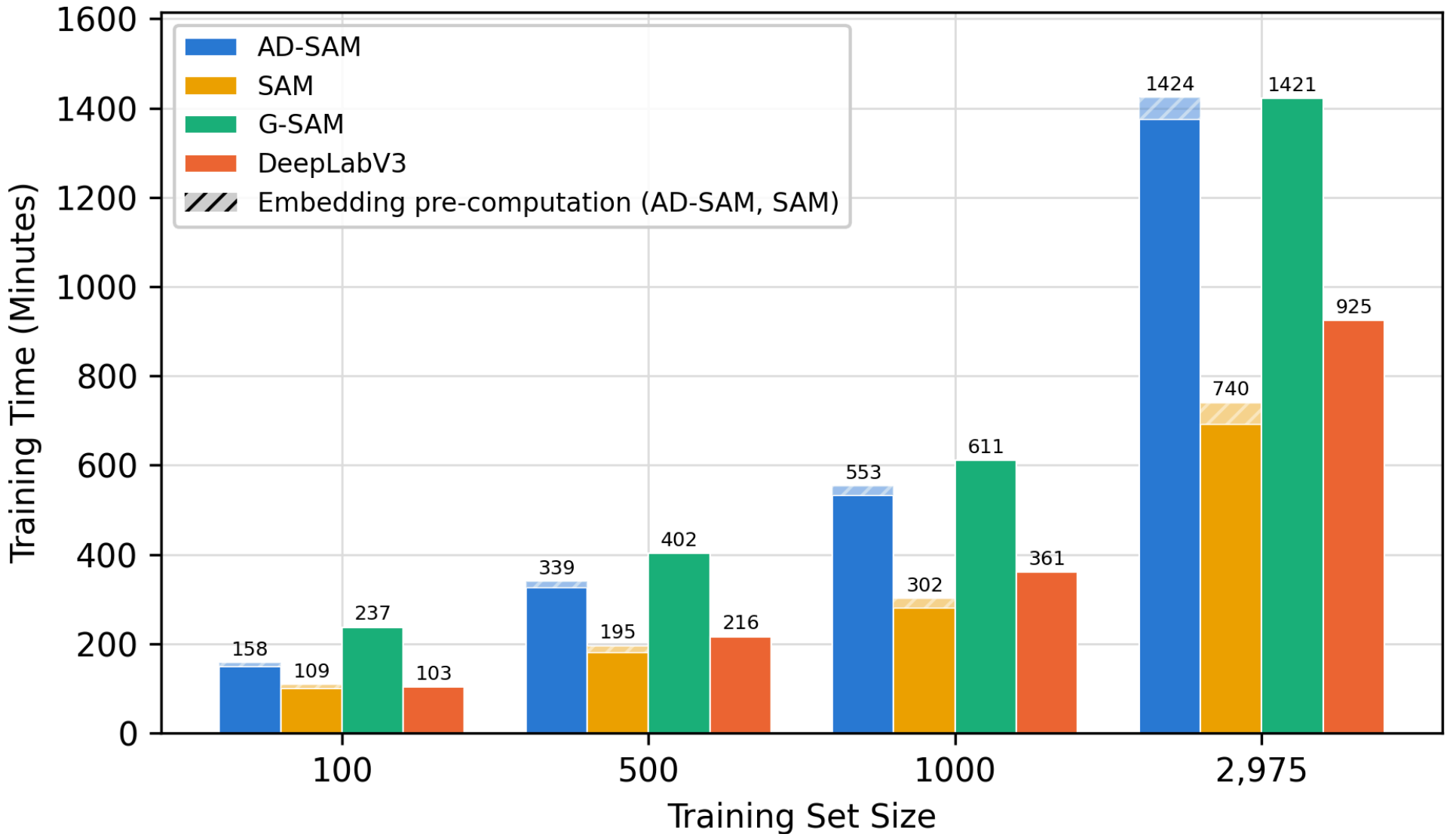

**(a) Cityscapes (full dataset)**

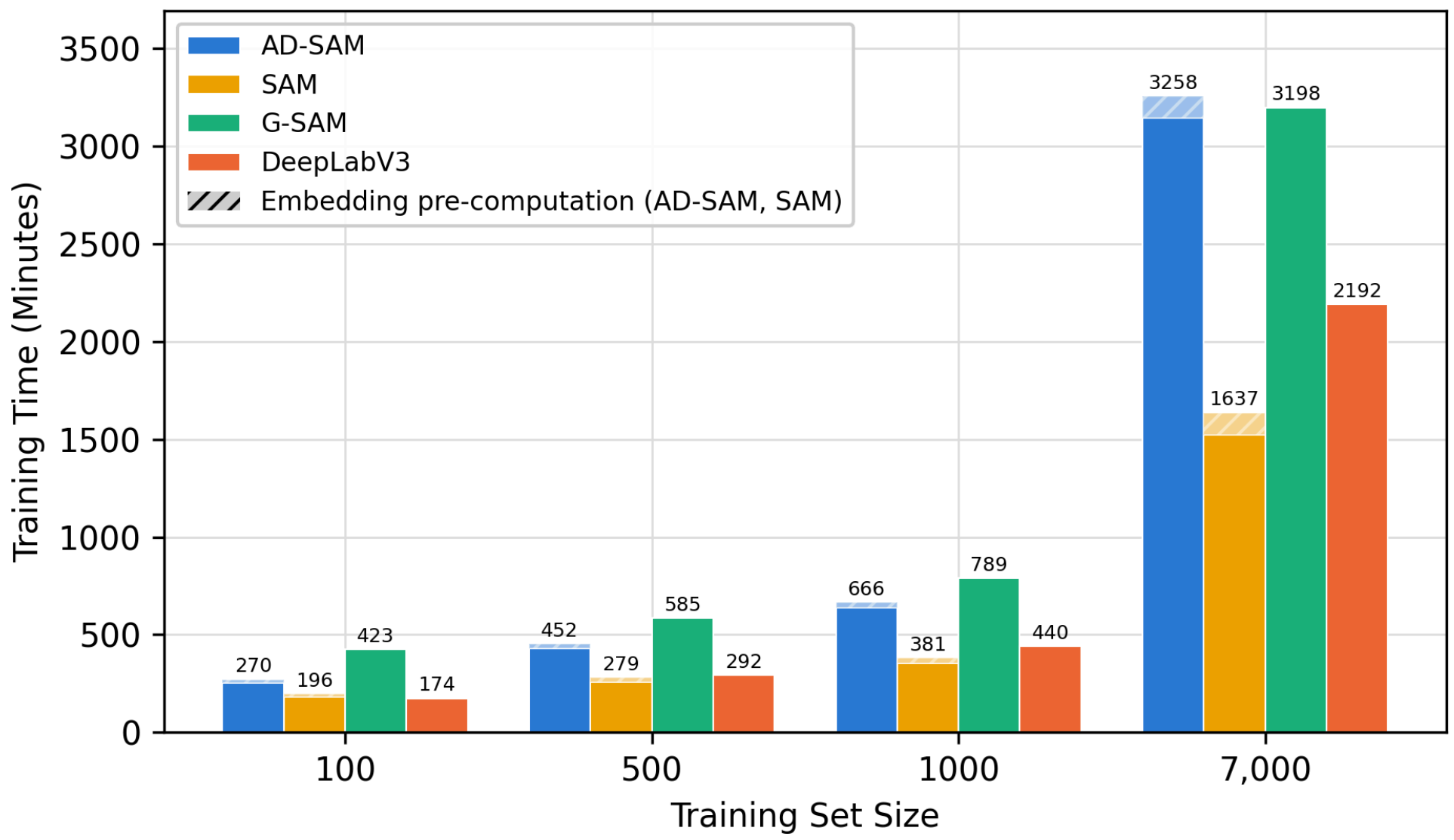

**(b) BDD100K (full dataset)**

**Figure 7. Training time across models and training-set sizes**

## 5.6.Per-Class Segmentation Analysis

Table 4 provides class-level results for AD-SAM across the 19 semantic categories. The model achieves high IoU for major structural and contextual classes on both datasets, including road (98.32% / 94.97%), building (92.63% / 87.56%), vegetation (92.24% / 87.91%), sky (94.81% / 95.72%), and car (94.87% / 90.92%) on Cityscapes / BDD100K, respectively.

These results indicate strong segmentation of prominent scene components that occupy substantial portions of urban driving images. Performance varies more substantially for smaller, less frequent, or visually challenging classes. Cityscapes yields higher IoU for several categories, including sidewalk (85.67% vs. 67.63%), wall (56.00% vs. 34.20%), traffic sign (79.27% vs. 63.86%), and bicycle (76.81% vs. 63.89%). However, this pattern is not universal, as BDD100K produces higher IoU for motorcycle (58.37% vs. 53.02%), and sky (95.72% vs. 94.81%). These differences highlight that class-level performance depends on the characteristics and representation of individual semantic categories within each dataset, rather than following a uniform dataset-level trend.

The most pronounced class-specific limitation occurs for train on BDD100K, for which AD-SAM obtains 0% IoU. This class appears in only 47 of the 7,000 training images and 7 of the 1,000 validation images (0.7%), and the model does not predict the class in the validation set. In contrast, train reaches 72.49% IoU on Cityscapes. This result illustrates the challenge posed by extremely sparse semantic categories motivating future investigation of class-aware sampling or training strategies for rare safety-relevant objects.

**Table 4. Class-wise segmentation performance (IoU in %) of AD-SAM across datasets**

| Class Label | Cityscapes | BDD100K |
|---|---|---|
| Road | 98.32 | 94.97 |
| Sidewalk | 85.67 | 67.63 |
| Building | 92.63 | 87.56 |
| Wall | 56.00 | 34.20 |
| Fence | 58.28 | 54.44 |
| Pole | 63.90 | 60.24 |
| Traffic light | 70.72 | 63.84 |
| Traffic sign | 79.27 | 63.86 |
| Vegetation | 92.24 | 87.91 |
| Terrain | 64.36 | 50.00 |
| Sky | 94.81 | 95.72 |
| Person | 81.27 | 77.08 |
| Rider | 63.60 | 44.32 |
| Car | 94.87 | 90.92 |
| Truck | 69.61 | 56.41 |
| Bus | 81.33 | 78.68 |
| Train | 72.49 | 0.00* |
| Motorcycle | 53.02 | 58.37 |
| Bicycle | 76.81 | 63.89 |
| **mIoU** | **76.27** | **64.74** |

**Note:** *Train appears in only 47 of the 7,000 BDD100K training images and 7 of the 1,000 validation images; AD-SAM does not predict this class in the validation set.

The qualitative examples in Figure 8 complement the class-level and sample-size analyses. On Cityscapes, predictions obtained with only 100 training images already capture much of the dominant scene structure, while increasing the sample size improves agreement with the ground truth and the representation of smaller scene elements. The effect of additional training data is more apparent on BDD100K, where predictions under the smallest data regime contain greater fragmentation and inaccuracies in complex scene regions, while larger training sets produce increasingly coherent segmentations. These visual patterns are consistent with the quantitative sample-size trends in Section 5.3 and further illustrate the greater sensitivity of AD-SAM to data availability on BDD100K.

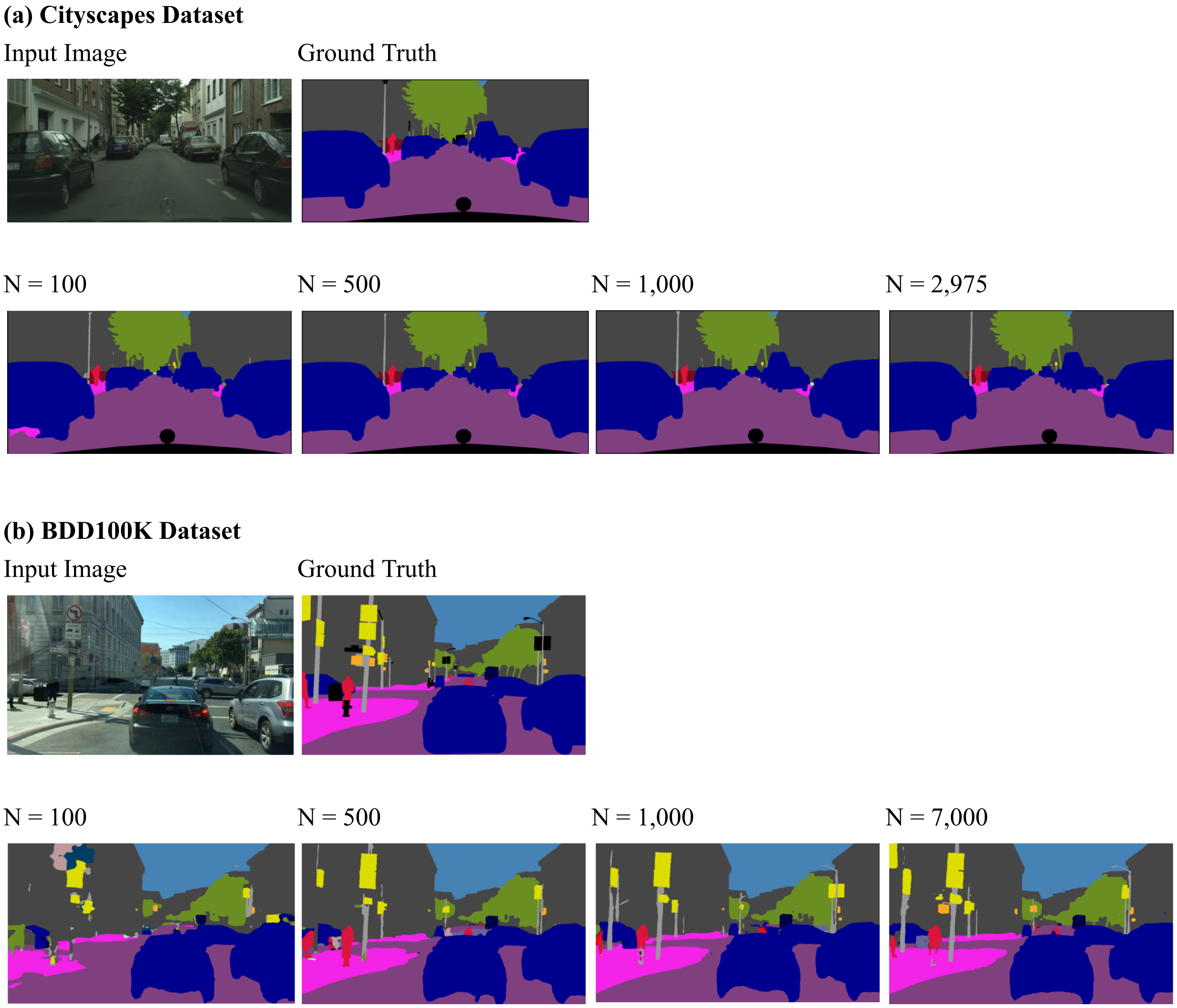

**Figure 8. Qualitative AD-SAM segmentation results across training-set sizes**

# 6. Conclusions

This study proposes AD-SAM, a SAM-based adaptation framework for semantic segmentation in autonomous driving. AD-SAM combines a frozen SAM encoder with a trainable multi-scale convolutional encoder, deformable feature fusion and decoding, and a hybrid loss formulation to integrate general visual representations with domain-specific features for prompt-free, closed-set semantic segmentation.

Experiments on Cityscapes and BDD100K show that AD-SAM achieves the highest overall performance among the evaluated models, reaching 76.27% and 64.74% mIoU, respectively. These results correspond to improvements of 3.45 and 5.00 percentage points over DeepLabV3, with larger gains over SAM and G-SAM. The sample-size analysis further reveals a dataset-dependent pattern. AD-SAM performs strongly across all training sizes on Cityscapes, whereas on the more heterogeneous BDD100K, its advantage emerges as more training data become available. Cross-dataset evaluation provides additional evidence of robustness to domain shift, with AD-SAM retaining 84.88% of its Cityscapes performance when evaluated directly on BDD100K, the highest retention among the evaluated models. The convergence analysis shows that AD-SAM reaches strong validation performance relatively early, while the computational analysis indicates an accuracy-cost trade-off. Its training time is comparable to G-SAM rather than lower than all baselines, while precomputing frozen SAM embeddings substantially reduces training memory requirements. Class-level and qualitative analyses further verify strong performance for major roadway and contextual categories while identifying remaining challenges for rare classes.

Several limitations provide directions for future research. First, performance remains class-dependent, particularly for extremely rare categories such as train in BDD100K, motivating class-aware sampling, loss weighting, or other strategies for imbalanced semantic classes. Second, although cross-dataset transfer was evaluated between Cityscapes and BDD100K, broader evaluation across geographic regions, sensing configurations, weather conditions, and additional driving datasets is needed to characterize domain robustness more comprehensively. Third, the present study focuses on image-based semantic segmentation. Future research can incorporate temporal information and complementary sensing modalities to improve robustness under occlusion and adverse conditions. Finally, although freezing and precomputing the SAM encoder reduces optimization memory, AD-SAM is not designed specifically for real-time inference. Future work can therefore investigate pruning, distillation, parameter-efficient adaptation, and lightweight architectures for deployment on resource-constrained platforms.

## CRediT Authorship Contribution Statement

**Het Patel:** Methodology, Data curation, Software, Formal analysis, Writing – original draft. **Mario Camarena:** Methodology, Data curation, Software, Formal analysis, Writing – original draft. **Fatemeh Nazari:** Conceptualization, Writing – original draft, Writing – review and editing, Funding acquisition. **Evangelos Papalexakis:** Conceptualization, Methodology, Formal analysis, Supervision, Funding acquisition. **Mohamadhossein Noruzoliaee:** Formal analysis, Writing – review and editing, Funding acquisition. **Jia Chen:** Conceptualization, Methodology, Formal analysis, Funding acquisition.

## Declaration of Competing Interest

The authors declare that they have no known competing financial interests or personal relationships that could have appeared to influence the work reported in this paper.

## Acknowledgements

This research was funded in part by the U.S. National Science Foundation (Award No. 2431569 and Award No. 2112650). The opinions expressed are solely those of the authors, and do not necessarily represent those of the U.S. National Science Foundation.

## Declaration of Generative AI Use

During the preparation of this work, the authors used OpenAI's ChatGPT to improve the clarity, organization, and readability of the manuscript. The tool was not used to generate, analyze, or interpret experimental data or research findings. After using this tool, the authors reviewed and edited the content as needed and take full responsibility for the content of the published article.

## Data Availability

Code is available at https://github.com/HettyPatel/AD-SAM. Cityscapes and BDD100K datasets are available from the respective providers.